\documentclass[10pt,twocolumn,letterpaper]{article}

\usepackage{cvpr}              %

\usepackage[dvipsnames]{xcolor}
\usepackage{paralist}
\usepackage{breqn}
\usepackage{tabularx}
\usepackage{amsmath}

\newcommand{\myparagraph}[1]{\vskip -8pt \noindent\textbf{#1}}

\newcommand{\suppmat}{Supplement}
\newcommand{\methodname}{Dream-in-4D}

\usepackage{tikz}
\usetikzlibrary{matrix}
\definecolor{cvprblue}{rgb}{0.21,0.49,0.74}
\usepackage[pagebackref,breaklinks,colorlinks,citecolor=cvprblue]{hyperref}

\title{A Unified Approach for Text- and Image-guided 4D Scene Generation}
\author{Yufeng Zheng$^{1,2,3}$, 
Xueting Li$^{1}$, 
Koki Nagano$^{1}$, 
Sifei Liu$^{1}$, 
Karsten Kreis$^{1}$, 
Otmar Hilliges$^{2}$, 
Shalini De Mello$^{1}$\\
$^{1}$NVIDIA, $^{2}$ETH Zurich, $^{3}$Max Planck Institute for Intelligent Systems
}
\begin{document}

\twocolumn[{
	\renewcommand\twocolumn[1][]{#1}%
	\maketitle
 	\begin{center}
    \includegraphics[trim=0em 0em 0em 0em, clip=true, width=\linewidth]{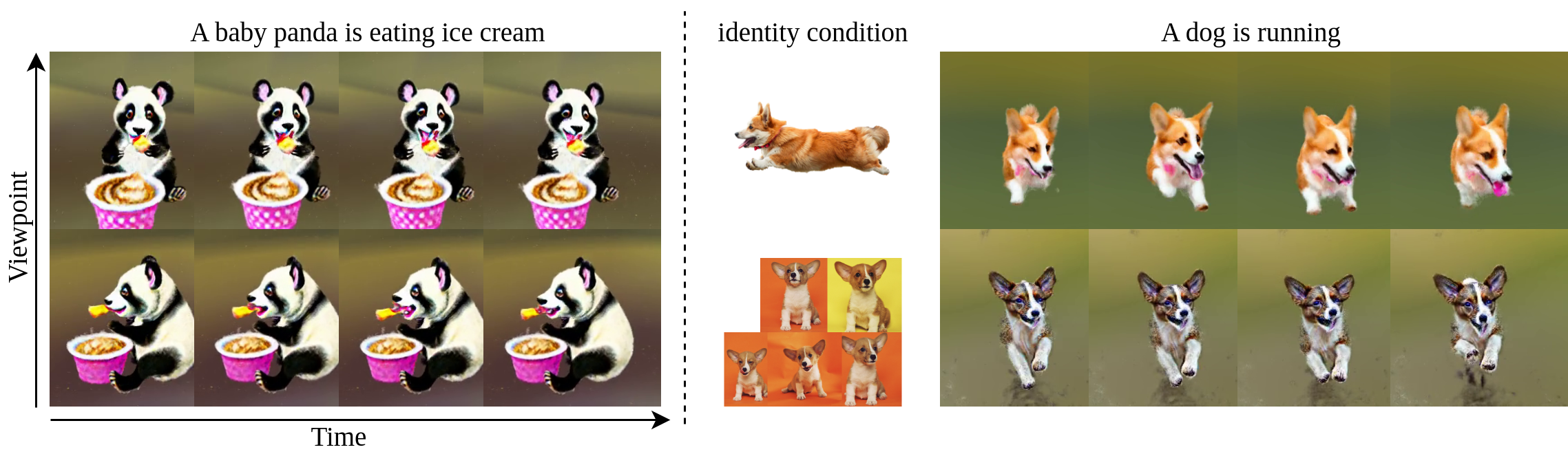}

    \vspace{-0.25cm}
    \captionof{figure}{\textbf{Text-to-4D. }Our method provides a unified approach for generating 4D dynamic content from a text prompt with diffuion guidance, supporting both unconstrained generation and controllable generation, where appearance is defined by one or multiple images.}
    \label{fig:teaser}
	\end{center}
}]
\maketitle
\begin{abstract}
\vspace{-0.4cm}
Large-scale diffusion generative models are greatly simplifying image, video and 3D asset creation from user-provided text prompts and images. However, the challenging problem of text-to-4D dynamic 3D scene generation with diffusion guidance remains largely unexplored. We propose \methodname, which features a novel two-stage approach for text-to-4D synthesis, leveraging
\begin{inparaenum}[(1)]
\item 3D and 2D diffusion guidance to effectively learn a high-quality static 3D asset in the first stage; 
\item a deformable neural radiance field that explicitly disentangles the learned static asset from its deformation, preserving quality during motion learning; and 
\item a multi-resolution feature grid for the deformation field with a displacement total variation loss to effectively learn motion with video diffusion guidance in the second stage. 
\end{inparaenum}
Through a user preference study, we demonstrate that our approach significantly advances image and motion quality, 3D consistency and text fidelity for text-to-4D generation compared to baseline approaches. 
Thanks to its motion-disentangled representation, \methodname{} can also be easily adapted for controllable generation where appearance is defined by one or multiple images, without the need to modify the motion learning stage. 
Thus, our method offers, for the first time, a unified approach for text-to-4D, image-to-4D and personalized 4D generation tasks.
\end{abstract}

\let\thefootnote\relax\footnotetext{
\vskip -1em

\noindent Work was partially done during an internship at NVIDIA. Project page:
\url{https://research.nvidia.com/labs/nxp/dream-in-4d/}}    
\vspace{-1cm}
\section{Introduction}
\label{sec:intro}

The advent of large-scale text-conditioned diffusion-based generative models for images has ushered in a new era of imaginative, high-quality image synthesis~\cite{rombach2021highresolution, deepfloydif, saharia2022photorealistic}. Their simple and intuitive conditioning in the form of text prompts are game-changers in democratizing visual content creation for non-expert users. Subsequently, these developments have also led to impressive progress in 
\begin{inparaenum}[(1)]
\item text- or image-conditioned static 3D content creation~\cite{poole2022dreamfusion, lin2023magic3d, metzer2023latent, qian2023magic123}, achieved by leveraging guidance from generic and 3D-aware~\cite{liu2023zero1to3, shi2023MVDream, liu2023syncdreamer} image diffusion models, and 
\item video content creation via video diffusion models~\cite{zeroscope,modelscope,blattmann2023videoldm, singer2022mav}.
\end{inparaenum}

However, for various real-world applications such as gaming, AR/VR, and advertising, synthesizing static 3D assets alone does not suffice. It is desirable to also animate 3D assets using intuitive user-provided text prompts to further save animators' time and level-of-expertise. To go beyond static 3D content creation, we delve into the largely unexplored problem of text-conditioned 4D scene generation, a.k.a., \emph{text-to-4D} synthesis with diffusion guidance. 
This is a challenging problem encompassing both text-to-3D and text-to-video synthesis. It requires learning not only a 3D-consistent representation of a static scene capable of free-view rendering, but also its plausible and semantically-correct dynamic 3D motion over time.

The pioneering work MAV3D~\cite{singer2023textto4d} is the first attempt to address this problem. It proposes a two-stage approach: the first to learn a static 3D asset, and the second to optimize its full dynamic representation with guidance from a video diffusion model~\cite{singer2022mav}. It further models the dynamic representation via a neural hexplane~\cite{cao2023hexplane}. While impressive in demonstrating feasibility, this early work leaves much room for improvement in terms of robustness, quality and realism. Additionally, it does not solve the problems of image-to-4D or personalized-4D content creation, wherein, in addition to a text prompt, an image or a set of images is provided as input to control the appearance of the 4D outcome.

\begin{figure}
     \centering
     \begin{subfigure}[b]{0.48\linewidth}
         \centering
         \includegraphics[width=\textwidth]{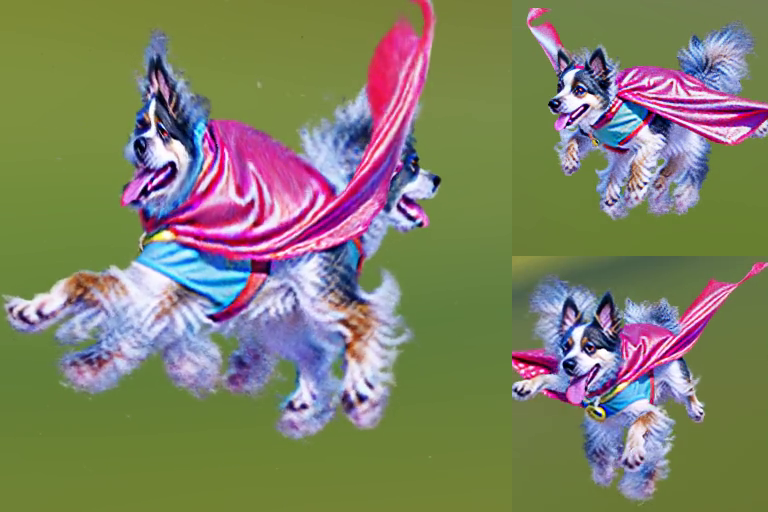}
         \caption{Multi-view images of a generated 3D asset without 3D diffusion guidance. It suffers from the Janus problem.}
         \label{fig:motivation_static}
     \end{subfigure}
     \hfill
     \begin{subfigure}[b]{0.50\linewidth}
         \centering
         \includegraphics[width=\textwidth]{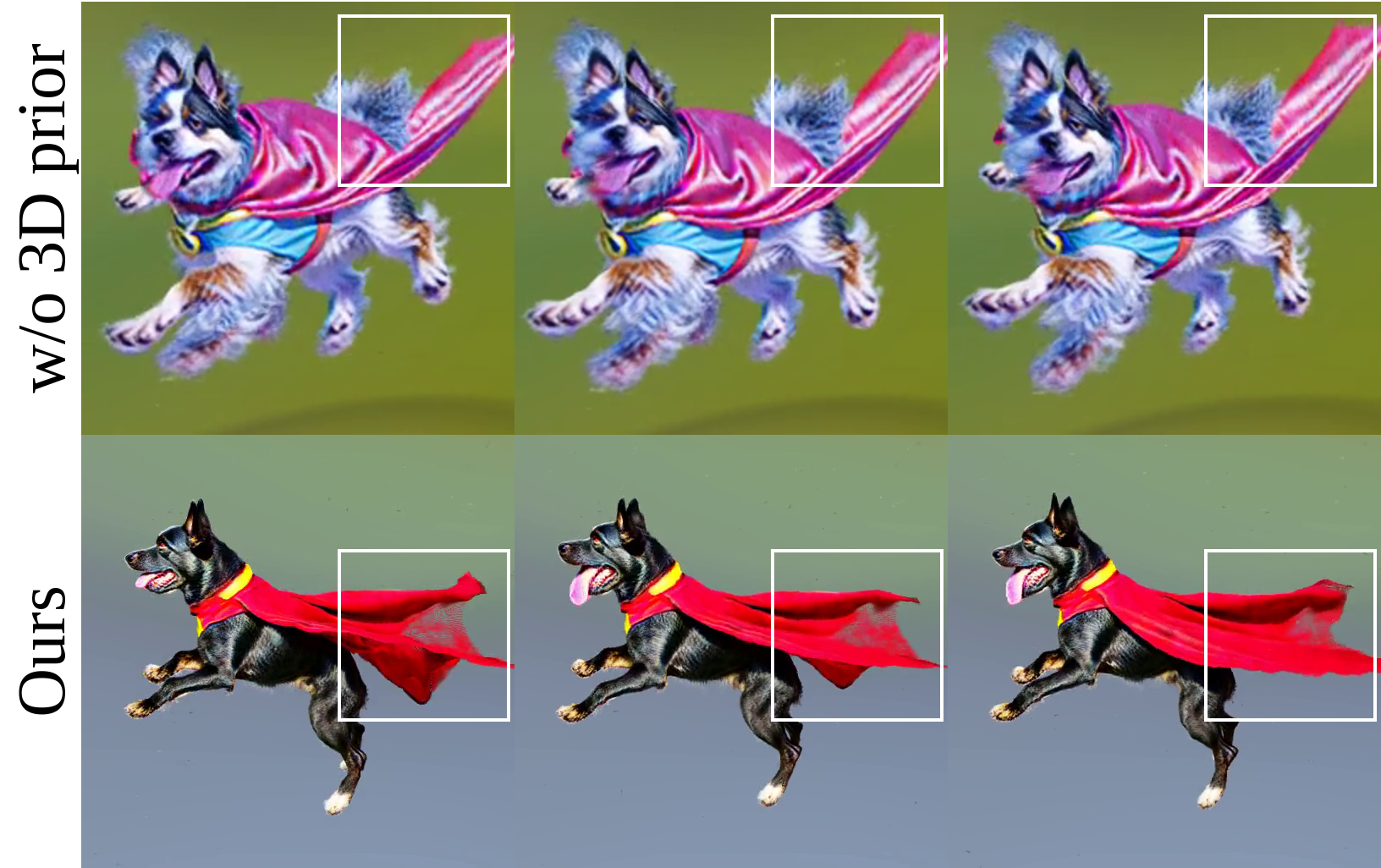}
         \caption{The presence of the Janus problem, further hinders learning of the cape's correct temporal motion (top row).}
         \label{fig:motivation_motion}
     \end{subfigure}
        \caption{We show the importance of a high-quality static asset for 4D content generation. The prompt for this scene is `A superhero dog with a red cape is flying through the sky'.}
        \label{fig:motivation}
        \vspace{-0.5cm}
\end{figure}
To address these challenges, we propose a novel method for text-to-4D dynamic scene synthesis, named \emph{\methodname}. It employs a two-stage approach, to first learn a static scene representation and then its motion. Our first primary insight is that achieving high-quality, 3D-consistent static reconstruction in the first stage is crucial for successfully learning motion in the second stage. For example, in Fig.~\ref{fig:motivation} we show a a multi-view \emph{inconsistent} 3D dog with two heads due to the Janus problem learned in the first stage. It introduces significant ambiguity for the second dynamic stage and substantially undermines the quality of the learned motion (in the dog's cape).
However, relying solely on guidance from image or video diffusion models for static text-to-3D synthesis, as proposed in~\cite{singer2023textto4d}, easily encounters the Janus problem (see the first row of Fig.~\ref{fig:wo_SD}). Therefore, we leverage 3D-aware~\cite{shi2023MVDream, liu2023zero1to3} and standard image diffusion models~\cite{rombach2021highresolution, deepfloydif} in stage-one to achieve high-quality view-consistent text-to-3D synthesis, along with video diffusion guidance~\cite{zeroscope} in stage-two to learn realistic motion. This forms our first key contribution, which is to leverage guidance from a carefully-designed combination of image, 3D, and video diffusion models to effectively solve the task of text-to-4D synthesis.

Our approach is further motivated by the observation that fine-tuning the static model with video diffusion models in stage-two leads to lower visual quality and prompt fidelity, primarily because these models are trained with lower quality videos compared to image diffusion models.  
To address this problem, our insight is to decompose the synthesis process into two \emph{distinct} training stages, the first of which is designed to learn a high-quality static 3D asset and the second dedicated to effectively animating it with the provided text prompt, while keeping the pre-trained static 3D asset unchanged. 
This, in turn, requires a 4D neural representation that fully disentangles the canonical static representation and its motion. However, this cannot be achieved with the hexplane~\cite{cao2023hexplane} representation proposed in~\cite{singer2023textto4d}, which entangles the static representation and its motion. To this end, we propose to use a variant of a deformable neural radiance field (D-NeRF)~\cite{pumarola2020dnerf} for the task of 4D content generation. %
A D-NeRF consists of a canonical 3D NeRF~\cite{mildenhall2021NeRF} and a 4D deformation MLP that maps time-dependent deformed space to the common canonical static space. With the proposed disentangled representation for 4D scene synthesis, we can freeze the pre-trained high-quality static 3D model from stage-one and only optimize the deformation field using video diffusion guidance in stage-two. %
To successfully learn detailed and realistic motion, we further encode the 4D deformation field with multi-resolution feature grids and regularize motion using a novel total variation loss on the rendered displacement maps. We find that the former enhances detailed motion while the latter reduces spatial and temporal jitter. 

Through a user preference study on diverse text prompts, we show that our algorithm achieves significant improvements in visual quality, 3D consistency, prompt matching and motion quality compared to alternate baselines. Furthermore, the ability to disentangle the canonical and motion representation allows for easy adaptation to image-conditioned 4D generation, without requiring modifications to the motion learning stage. Thus, we demonstrate image-to-4D generation given a single-view image, and personalized 4D generation using 4-6 casually captured images of a subject (See Fig.~\ref{fig:teaser}) with our unified \emph{\methodname} method.

In summary, our key contributions include:
\begin{compactenum}
    \item We propose to combine image, 3D-aware and video diffusion priors for the text-to-4D task, significantly improving the visual quality, 3D consistency and text-fidelity of the learned static assets in the first stage.
    \item By explicitly disentangling the static representation from its deformation, our method preserves the high-quality static asset during motion learning.
    \item We propose to use a multi-resolution feature grid and a total variation loss on the deformation field to effectively learn motion with video diffusion guidance.
    \item We demonstrate that our method offers, for the first time, a unified approach for text-to-4D, image-to-4D and personalized 4D generation tasks.
\end{compactenum}

\section{Related Work}
\label{sec:related}
\myparagraph{Dynamic neural radiance fields.}
Modeling dynamic 3D content with NeRFs has been extensively studied in the novel-view synthesis literature. To extend NeRFs to dynamic scene modeling, previous works either learn a high-dimensional radiance field conditioned on temporal embeddings~\cite{martin2021nerf, li2022neural}, or a separate deformation mapping to model motion~\cite{pumarola2020dnerf, park2021nerfies}. To speed up training and inference, plane- and voxel-based feature grids are combined with MLPs to formulate efficient hybrid NeRF representations~\cite{muller2022ingp, fridovich2022plenoxels, chan2022efficient}, which are extended to dynamic scene modeling by learning additional planes for the temporal dimension~\cite{fridovich2023kplane, cao2023hexplane}. In this work, we leverage a deformable NeRF representation where the canonical geometry and 4D deformation field are both encoded by multi-resolution feature grids~\cite{muller2022ingp}. As a result, the geometry and motion are fully disentangled, which not only eases motion learning but also allows easy adaptation to various applications such as image(s)-to-4D video generation.
\\
\myparagraph{Diffusion models.}
Recently, diffusion models have revolutionized the computer vision community by showing remarkable advancements in image, video or novel view image synthesis. Seminal works such as Stable Diffusion~\cite{rombach2021highresolution} and DeepFloyd~\cite{deepfloydif} take a text prompt as input and produce high-quality images that align with the prompt. Leveraging large-scale image datasets, these diffusion models learn various prior knowledge ranging from object appearance to complex scene layout. One line of subsequent works~\cite{zeroscope,modelscope,blattmann2023videoldm} fine-tunes text-to-image diffusion models on video datasets, successfully extending them to generate realistic videos matching both the object and motion described by the input prompt. Another line of methods~\cite{liu2023zero1to3,liu2023syncdreamer} learns 3D-aware diffusion models with images rendered from synthetic objects~\cite{objaverse,objaverseXL}. By conditioning on camera parameters, these methods produce novel view images of an object that are consistent with each other and align with the observed view.
\\
\myparagraph{Text/image(s) to 3D with diffusion priors.}
Beyond direct sampling from the diffusion models, several works employ image diffusion priors as an optimization signal for 3D generation. The pioneering work, DreamFusion~\cite{poole2022dreamfusion}, optimizes a 3D model by presenting its renderings to a text-to-image diffusion model and acquiring gradient supervision through Score Distillation Sampling (SDS). Subsequent works enhance the synthesis quality and speed by incorporating the mesh representation~\cite{lin2023magic3d, tsalicoglou2023textmesh}, advancing score distillation~\cite{wang2023prolificdreamer,zhu2023hifa,sjc}, exploring representations in the latent space~\cite{metzer2023latent}, or disentangling geometry and texture~\cite{Chen_2023_ICCV}. Yet, these methods suffer from the Janus problem due to the lack of 3D prior in the text-to-image diffusion models. To overcome this limitation, several works~\cite{shi2023MVDream,liu2023zero1to3, liu2023syncdreamer, threestudio2023} leverage 3D-aware diffusion models as supervisions, generating 3D objects that are consistent across multiple views. In addition to text, a few approaches further take one or multiple images as an input and reconstruct a 3D object matching both the prompt and the image(s). The former~\cite{tang2023makeit3d,melaskyriazi2023realfusion,qian2023magic123} simultaneously utilize text-to-image and novel view diffusion models~\cite{liu2023zero1to3}, while the latter~\cite{shi2023MVDream,raj2023dreambooth3d} overfit a diffusion model on a few images depicting the same subject to achieve personalized diffusion guidance.
\\
\myparagraph{Text-to-4D with diffusion priors.} In this paper, we go beyond 2D/3D generation and aim to synthesize a 4D video given a text prompt. This is hitherto a highly challenging and under-explored domain. The most relevant work to ours is MAV3D~\cite{singer2023textto4d}, which optimizes a 4D scene by leveraging a pre-trained video diffusion model. However, due to the entanglement of geometry and motion, as well as a lack of a 3D-aware prior, this method suffers from the Janus problem and produces low quality texture. To resolve these limitations, we leverage a carefully designed combination of image, video and 3D-aware diffusion models and fully disentangle the geometry and motion. Our method synthesizes multi-view consistent 4D videos with realistic appearance and motion. Furthermore, the disentanglement of canonical and motion representations readily enables novel applications such as image-to-4D and personalized 4D generation.

\section{Method}
\begin{figure*}[t]
\begin{center}
\includegraphics[trim=5em 0em 0em 0em, clip=true, width=\linewidth]{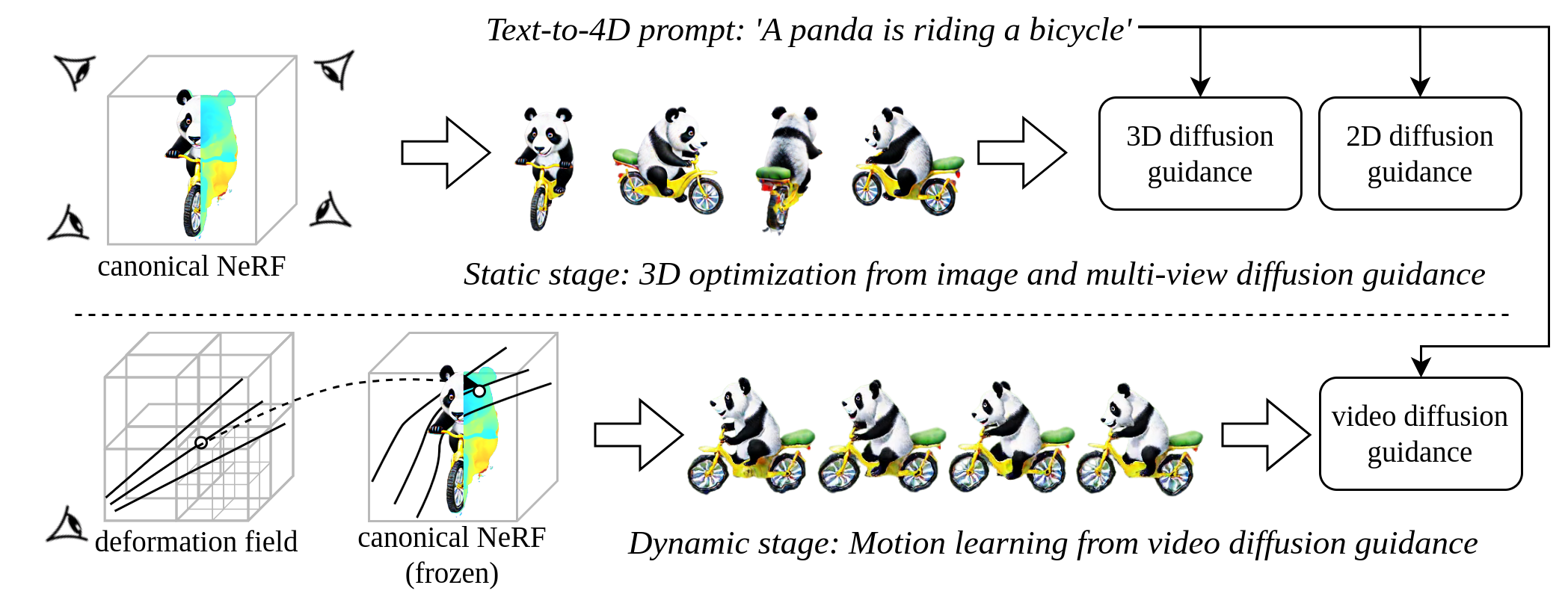}
\end{center}
\vspace{-0.5cm}
\caption{
\textbf{Method overview.} Adopting a two-stage approach, \emph{\methodname{}} first utilizes 3D and 2D diffusion guidance to learn a static 3D asset based on the provided text prompt (top). Then, it optimizes a deformation field using video diffusion guidance to model the motion described in the text prompt (bottom). Featuring a motion-disentangled D-NeRF representation, our method freezes the pre-trained static canonical asset while optimizing for the motion, achieving high quality view-consistent 4D dynamic content with realistic motion.}
\label{fig:pipeline}
\end{figure*}
Given a text prompt and optionally one or a few images to specify the object's appearance, we aim to generate a 4D video that matches both the object and the motion described in the prompt. To this end, we propose a two-stage training pipeline. In the first static stage (Sec. ~\ref{sec:static_stage}), we synthesize a high-quality static 3D scene using both 2D and 3D diffusion priors. In the second dynamic stage (Sec. ~\ref{sec:dynamic_stage}), we learn the 3D motion of the scene using a video diffusion model, while keeping the static scene representation intact.

\subsection{Static Stage}
\label{sec:static_stage}
The goal of the static stage is to generate a high-quality 3D scene that aligns with the text prompt. 
To efficiently learn a canonical model of a 3D scene, we opt for the NeRF~\cite{mildenhall2021NeRF} representation with multi-resolution hash-encoded features~\cite{muller2022ingp}, which is extensively used in previous text-to-3D methods~\cite{poole2022dreamfusion, lin2023magic3d, wang2023prolificdreamer, zhu2023hifa, metzer2023latent}. This static 3D model is accompanied by a deformation field to represent the dynamic motion of the 3D scene in the subsequent motion learning stage.
Two elements of the static stage are crucial for 3D asset quality and play an important role in facilitating motion learning in the subsequent dynamic stage: the generated 3D object(s)
\begin{inparaenum}[(1)] %
\item should be view-consistent (\textit{i.e.}, free of the Janus problem) and 
\item should follow the spatial composition described in the text prompt. 
\end{inparaenum}
Intuitively, the former reduces contradictory gradients from different views in deformation optimization while the latter eases motion learning by presenting a reasonable spatial layout of multiple objects. 
For instance, the panda sitting on top of a bike in Fig.~\ref{fig:pipeline} sets a good starting point for the dynamic stage to learn the ``riding'' motion. To achieve these goals, we propose to utilize both 3D and generic 2D diffusion models for the static stage. This is, in spirit, similar to prior work~\cite{qian2023magic123} for image-to-3D synthesis.
In the following, we introduce the 3D and 2D diffusion guidance used for stage-one. 

\begin{figure}
\centering
\newcommand{\myheight}{1.7cm}
\setlength\tabcolsep{1pt}
\begin{tikzpicture}[every node/.style={inner sep=0,outer sep=0}]
\matrix[matrix of nodes, nodes={anchor=center}, column sep=1mm, row sep=1mm]{
\rotatebox[origin=t]{90}{\footnotesize w/o MVDream}
&\includegraphics[height=\myheight]{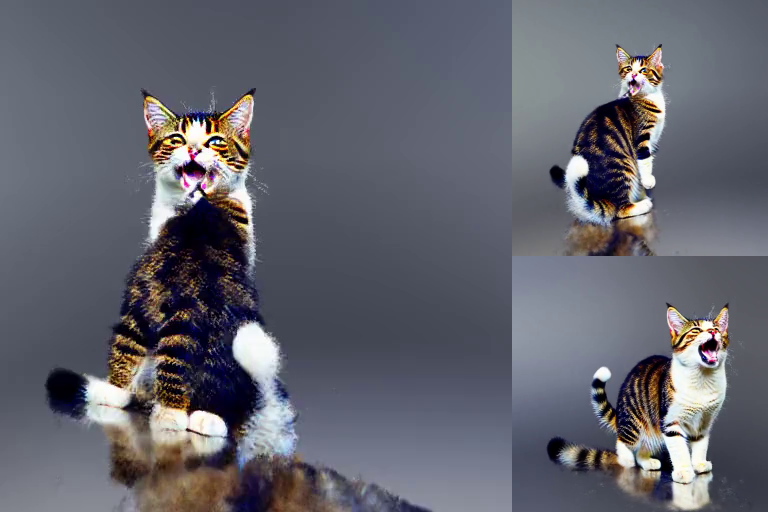}
&\includegraphics[height=\myheight]{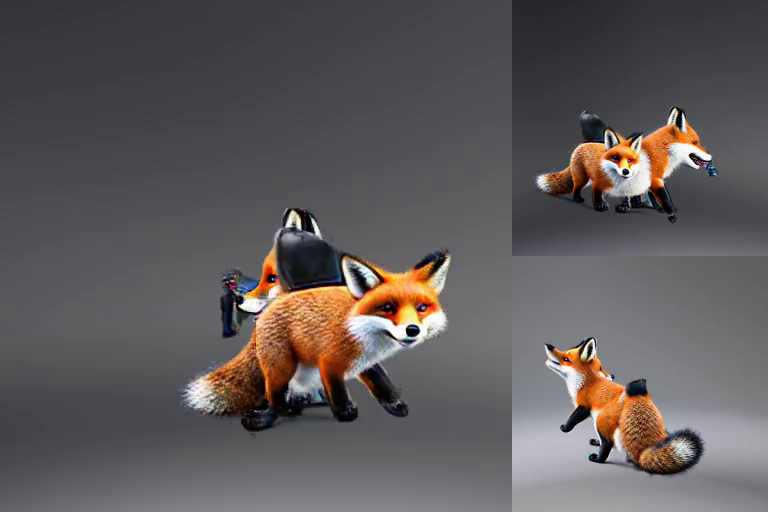}
&\includegraphics[height=\myheight]{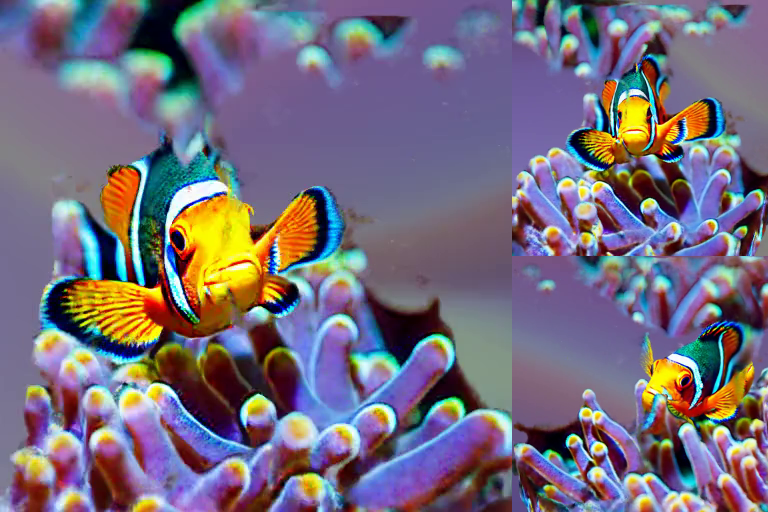}
\\
\rotatebox[origin=t]{90}{\footnotesize w/o SD}
&\includegraphics[height=\myheight]{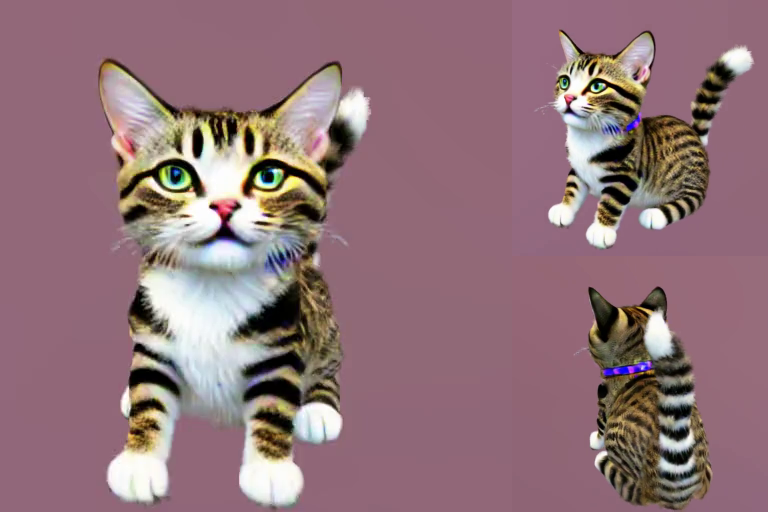}
&\includegraphics[height=\myheight]{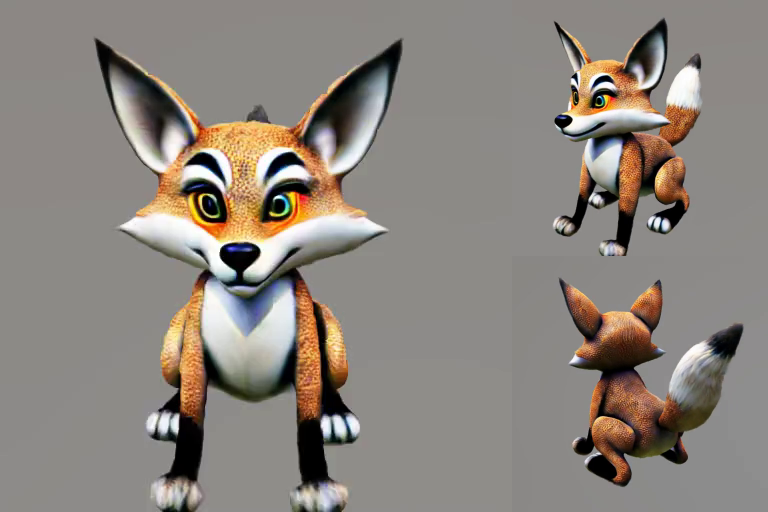}
&\includegraphics[height=\myheight]{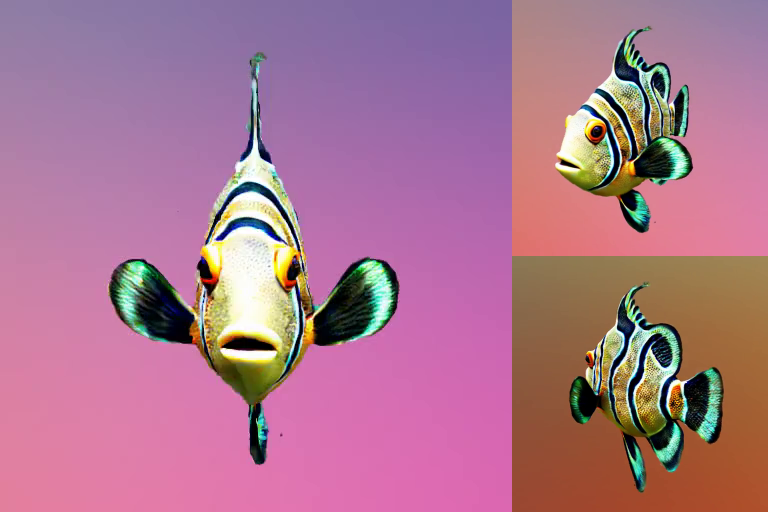}
\\
\rotatebox[origin=t]{90}{\footnotesize w/o Ours}
&\includegraphics[height=\myheight]{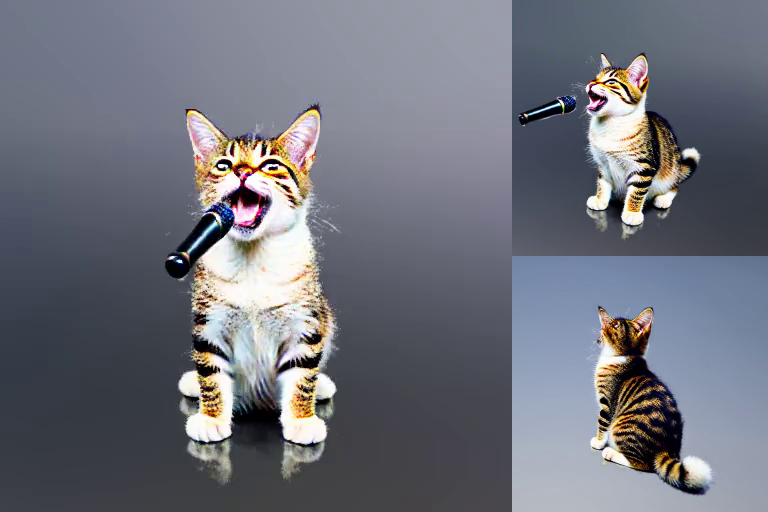}
&\includegraphics[height=\myheight]{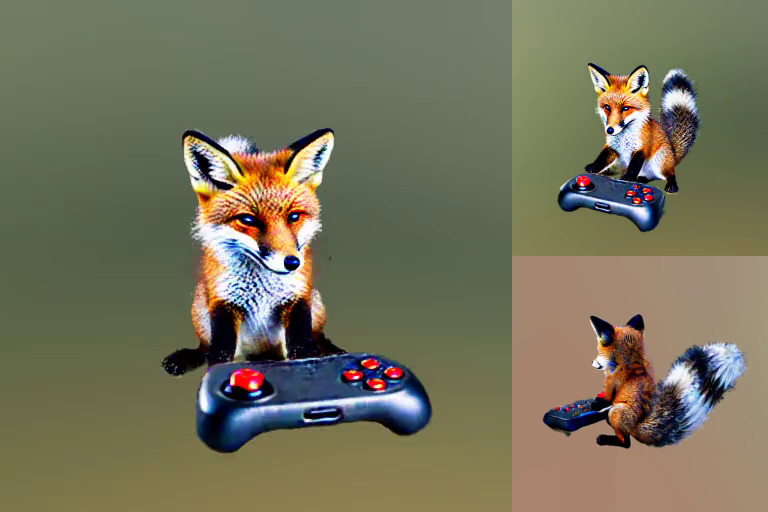}
&\includegraphics[height=\myheight]{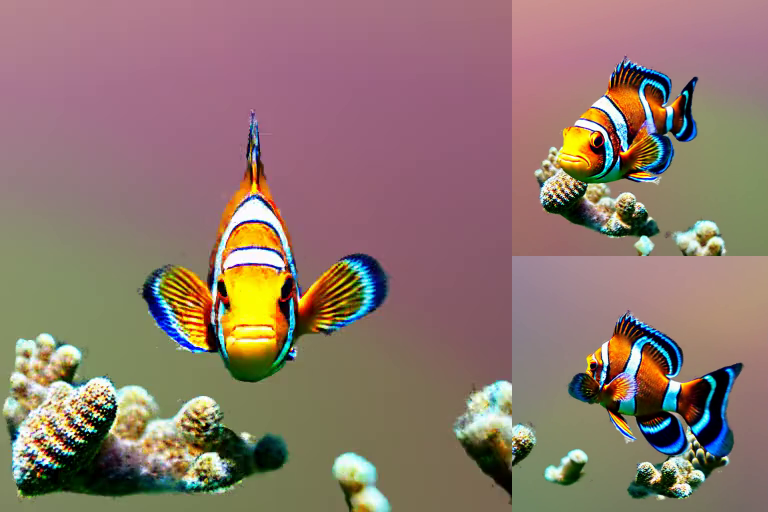}
\\
 &\scriptsize{A cat singing} & \scriptsize{A fox playing video game} & \scriptsize{Clown fish in coral reef}\\
};
\end{tikzpicture}
\caption{
\textbf{Static stage. }Without StableDiffusion guidance, the learned static model fails to learn the correct composition. Without MVDream guidance, the learned assets suffer from the Janus problem and contain multiple faces. Using guidance from both StableDiffusion and MVDream, results in the best text prompt fidelity and 3D consistency.
\label{fig:wo_SD}
}
\end{figure}

3D diffusion models~\cite{shi2023MVDream, liu2023zero1to3, liu2023syncdreamer} take camera parameters with a text prompt or an image as inputs and synthesize novel view images of the target object. Fine-tuned from image diffusion models with rendered images of synthetic 3D data~\cite{deitke2023objaverse}, 3D diffusion models provide valuable prior knowledge of the 3D world and enforce different views of a 3D object to be consistent. For text to 3D generation, we adopt the MVDream~\cite{shi2023MVDream} model to provide a 3D prior. Specifically, we render the synthesized 3D object from four different viewpoints (\textit{i.e.}, front, back, and two side views) and obtain guidance from a pre-trained MVDream model through the SDS loss~\cite{poole2022dreamfusion}. We use a reconstruction formulation of the SDS loss, similar to \cite{shi2023MVDream}. The 3D guidance loss is denoted as $\mathcal{L}_{3D}(I)$. 

However, due to the limited scale and synthetic nature of the 3D training datasets, static NeRF models optimized using MVDream alone tend to have synthetic-looking texture~\cite{shi2023MVDream}, and occasionally fail to produce realistic scene layouts 
(see the second row of Fig.~\ref{fig:wo_SD} where objects in the prompt are missing from the scene). 
Meanwhile, we observe that image diffusion models trained with large-scale 2D images encourage both realistic appearance and reasonable scene layouts, but by themselves, easily suffer from the Janus problem (see the first row of Fig.~\ref{fig:wo_SD}). Thus, we propose to combine 2D diffusion guidance with 3D guidance in stage-one. Specifically, we use StableDiffusion-v2.1 with the SDS objective (denoted as $\mathcal{L}_{2D}(I)$) to provide 2D guidance. The overall objective for stage-one is:
\begin{align*}
    \mathcal{L} = \lambda_{2D}\mathcal{L}_{2D}(I) + \lambda_{3D}\mathcal{L}_{3D}(I), %
\end{align*}
where $I$ denotes the set of rendered images from the sampled camera viewpoints, and $\lambda_{2D/3D}$ are the weights for the 2D and 3D guidances (see \suppmat{}
for the loss weights).

As shown in Fig.~\ref{fig:wo_SD}, by combining both the 2D and 3D diffusion models for guidance, our static stage generates 3D-consistent object(s) with realistic texture and plausible scene layouts.
\subsection{Dynamic Stage}
\label{sec:dynamic_stage}
\begin{figure}
\begin{center}
\newcommand{\myheight}{1.8cm}
\setlength\tabcolsep{1pt}
\begin{tabularx}{\textwidth}{ccc}
\includegraphics[height=\myheight]{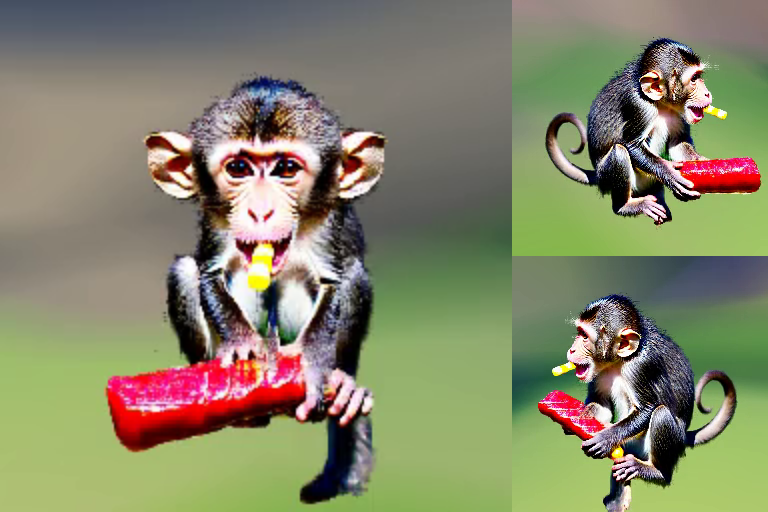}
&\includegraphics[height=\myheight]{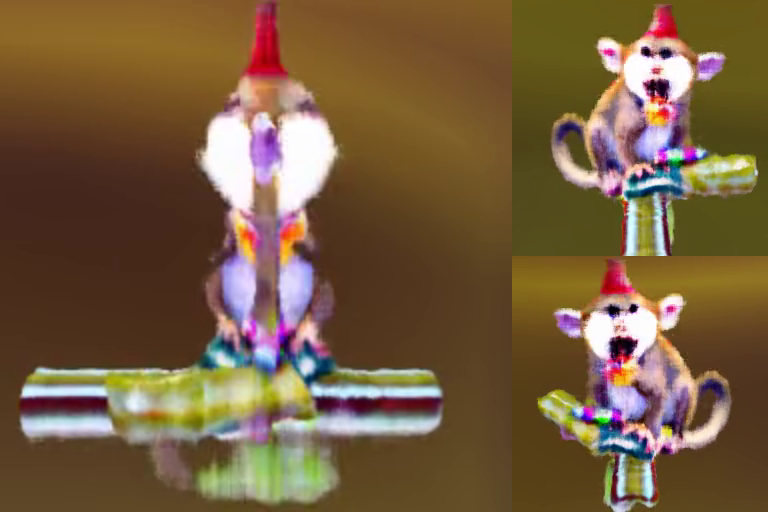}
&\includegraphics[height=\myheight]{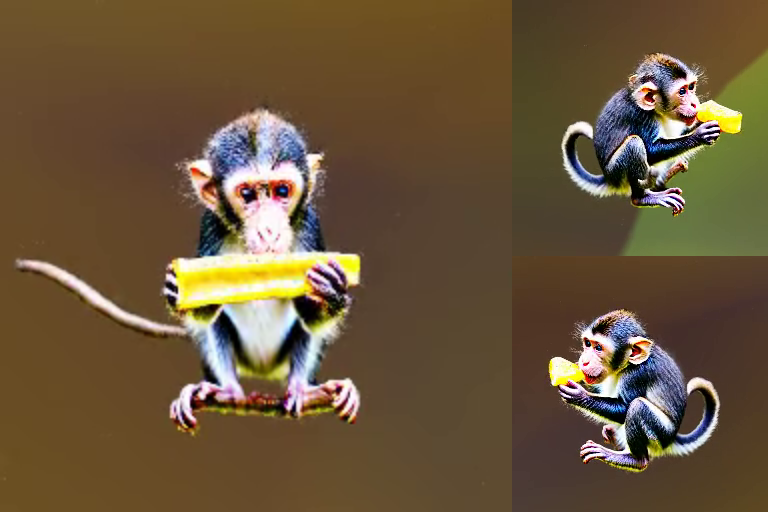}
\\
\small{Hexplane static} & \small{Hexplane dynamic} & \small{Ours dynamic}
\end{tabularx}
\caption{
\textbf{Hexplane v.s. deformable NeRF. } With a hexplane representation, even though the static stage successfully learns a high-quality 3D asset (column 1), its motion learning stage with video diffusion guidance still leads to degradation in texture and re-appearance of the Janus problem (column 2). 
\label{fig:hexplane}
}
\end{center}
\vspace{-0.75cm}
\end{figure}
\begin{figure}[t]
\begin{center}
\includegraphics[trim=0em 0em 0em 0em, clip=true, width=1.0\linewidth]{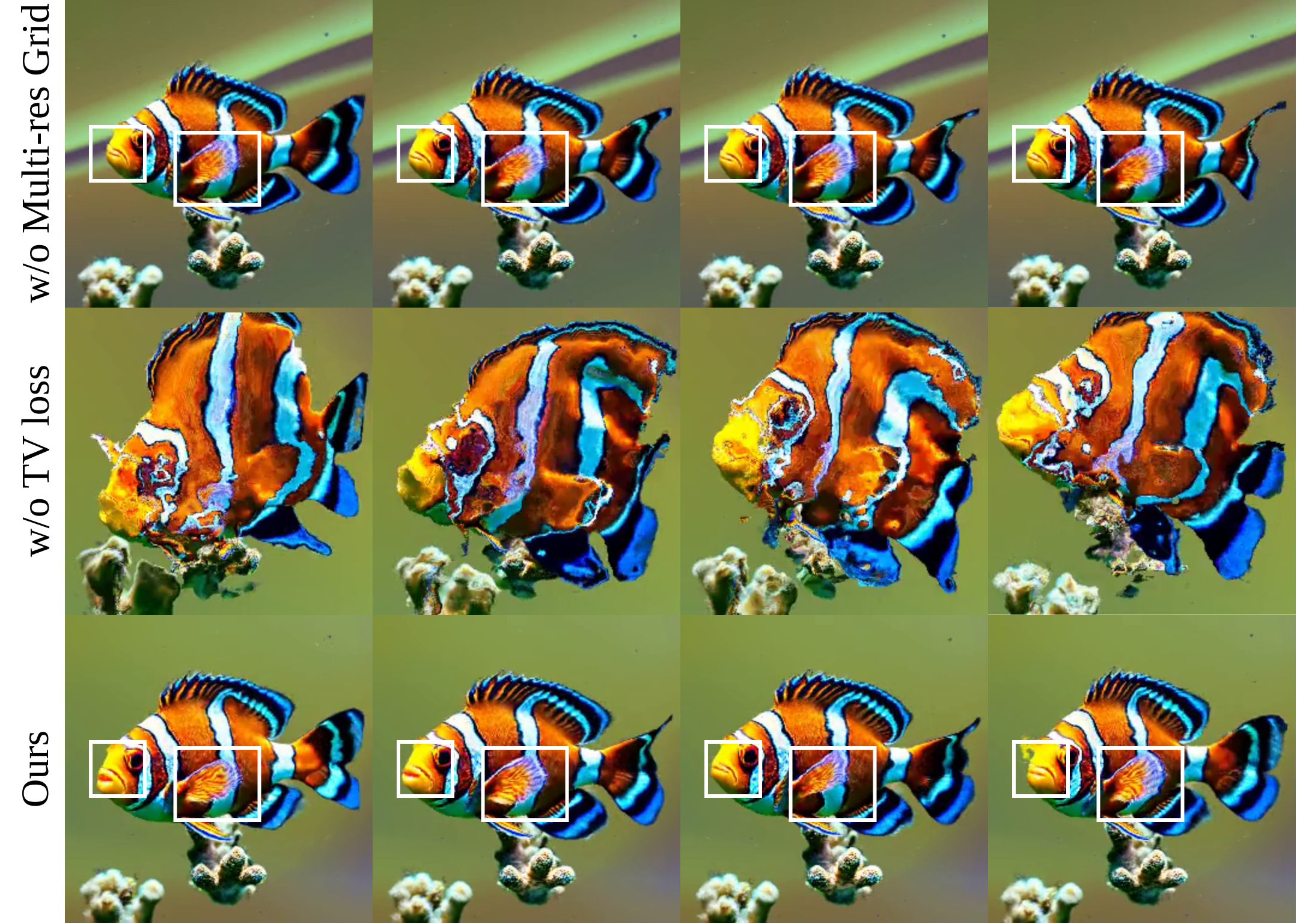}
\end{center}
\caption{
\textbf{Deformation learning. }The deformation MLP equipped with positional encoding instead of multi-resolution feature grids cannot capture local motions (mouth and fin in row 1). Without the proposed total variation loss on the displacement, the learned deformation contains substantial noise (row 2). Our approach, with both, results in the best quality (row 3).}
\label{fig:motion_ablation}
\vskip -1em
\end{figure}

In the dynamic stage, our goal is to learn a deformation field that animates the 3D scene generated in the static stage using guidance from a video diffusion model. As aforementioned, our key observation is that although video diffusion models provide a valuable motion prior, they are not 3D-aware and 
tend to produce unappealing visual results (see Fig.~\ref{fig:hexplane}, column 2). Therefore, we propose to fully disentangle the static model and the motion by freezing the NeRF network learned in the static stage and only learn the deformation field to match the motion described in the text prompt in the dynamic stage. Such a design brings two advantages: 
\begin{inparaenum}[(1)] %
\item it preserves the view consistency and high-quality texture learned in the static stage and 
\item it readily enables applications such as image-to-4D and personalized 4D generation (see Sec.~\ref{sec:image_conditioned_generation}).
\end{inparaenum}
\\
\myparagraph{Motion-disentangled 4D representation. } Our dynamic 4D representation consists of a canonical 3D radiance field (as described in Sec.~\ref{sec:static_stage}) and a deformation field. 
The deformation field is a 4D to 3D time-dependent mapping
$D(\mathbf{x_d}, t) \rightarrow \mathbf{x_c}$, where $\mathbf{x_d}$ is a 3D point's location in deformed space at time $t$, and $\mathbf{x_c}$ is its corresponding canonical location. 
Our insight is that the deformable field should be smooth both spatially and temporally due to the limited elasticity and velocity of the object. As a result, the deformation field does not require as high-resolution a feature grid as its static canonical 3D counterpart. Therefore, we utilize a 4D multi-resolution hash-encoded feature grid with a maximum resolution of 232 for the deformation field, in contrast to the maximum resolution of 4096 for the canonical static NeRF representation. Additionally, we found the usage of multi-resolution features to be crucial for learning correct local motion (see Fig.~\ref{fig:motion_ablation}). 
\\
\myparagraph{Motion optimization with video diffusion models. }
The deformation field is optimized via score distillation sampling using a video diffusion model. Specifically, 
we sample a static camera parameter, and render a 24-frame video $\mathbf{V}$ from our 4D representation. The time stamps are sampled evenly and the length of the video is randomly chosen between 0.8 and 1 (assuming that the full length is 1). 
We leverage a variant of the SDS loss~\cite{zhu2023hifa} for video diffusion guidance, where we predict the original video with 1-step denoising, and use a combination of a latent feature loss and a decoded RGB space loss. The video diffusion guidance loss can be expressed as $\mathcal{L}_{video}(\mathbf{V})  = \mathcal{L}_{latent}(\mathbf{V}) + 
\lambda_{dec}\mathcal{L}_{dec}(\mathbf{V}) 
$, where $\lambda_{dec}=0.1$.
We choose to use the Zeroscope~\cite{zeroscope} video diffusion model as our motion prior, but our method is robust to other models such as Modelscope~\cite{modelscope} (see our project webpage for results).
We found that matching the resolution of the learned videos to that of the video diffusion models to be important for successfully distilling motion priors. For Zeroscope, we render videos at a resolution of $144\times72$ and upsample them to $576\times320$ when training with video diffusion guidance. 
\\
\myparagraph{Total variation motion regularization. }To reduce temporal and spatial jitter in motion, we propose to use a novel total variation loss for the learned deformation (see Fig.~\ref{fig:motion_ablation}). Specifically, in addition to the RGB video $\mathbf{V}$, we also render a video of 3D displacements $\mathbf{D}$. The total variation loss on the rendered displacement video $\mathbf{D}$ can be expressed as:
\begin{align*}
\mathcal{L}_{TV}(D)=\sum_{x,y,t}{}{}(
||\mathbf{D}_{x-1, y,t} - \mathbf{D}_{x, y,t}||_2^2 \\
+ ||\mathbf{D}_{x, y-1,t} - \mathbf{D}_{x, y,t}||_2^2 
+ ||\mathbf{D}_{x, y,t-1} - \mathbf{D}_{x, y,t}||_2^2 ).
\end{align*}
The overall objective function for the second stage is then:
\begin{align*}
\mathcal{L} = \mathcal{L}_{video}(\mathbf{V}) + 
\lambda_{TV}\mathcal{L}_{TV}(\mathbf{D}),
\end{align*}
where $\lambda_{TV}=1000$.

\subsection{4D Generation Given One or Multiple Images}
\label{sec:image_conditioned_generation}
While text-to-4D generation is useful for many scenarios, there is a common desire to create content that features a specific object. However, language alone may be insufficient to describe the unique appearance of a given object. 
Thanks to the full disentanglement of its static and dynamic parts, our method can be easily extended to image-guided 4D generation, without modifying the motion learning stage.
In the following, we show that this can be done by simply replacing the diffusion models used in the static stage of our method.
\\
\myparagraph{Image-to-4D generation. } Given a single image, we reconstruct the corresponding 3D asset by replacing MVDream with an image-conditioned 3D diffusion model. Specifically, we use zero123-xl~\cite{liu2023zero1to3} as our 3D diffusion model and DeepfloydIF~\cite{deepfloydif} as our 2D diffusion model. Additionally, we supervise the reference view with the given image and its estimated foreground mask, similarly to \cite{qian2023magic123, liu2023zero1to3}. Fig.~\ref{fig:image-to-4D} shows examples of synthesized 4D videos from a single-view image.
\\
\myparagraph{Personalized 4D generation. } Given a few casually captured images of an object, Dreambooth~\cite{ruiz2023dreambooth} finetunes image diffusion models to generate personalized images of this object given a text prompt. By replacing our generic image diffusion model with a finetuned, personalized version, we can create personalized 3D assets given a text prompt and a few casual images. Specifically, we use personalized StableDiffusion together with MVDream for this task. We show synthesized 4D videos in Fig.~\ref{fig:personalized_4D}.
\begin{figure*}
\begin{center}
\newcommand{\myheight}{4.5cm}
\setlength\tabcolsep{1pt}
\begin{tabularx}{\textwidth}{cc}

\includegraphics[height=\myheight]{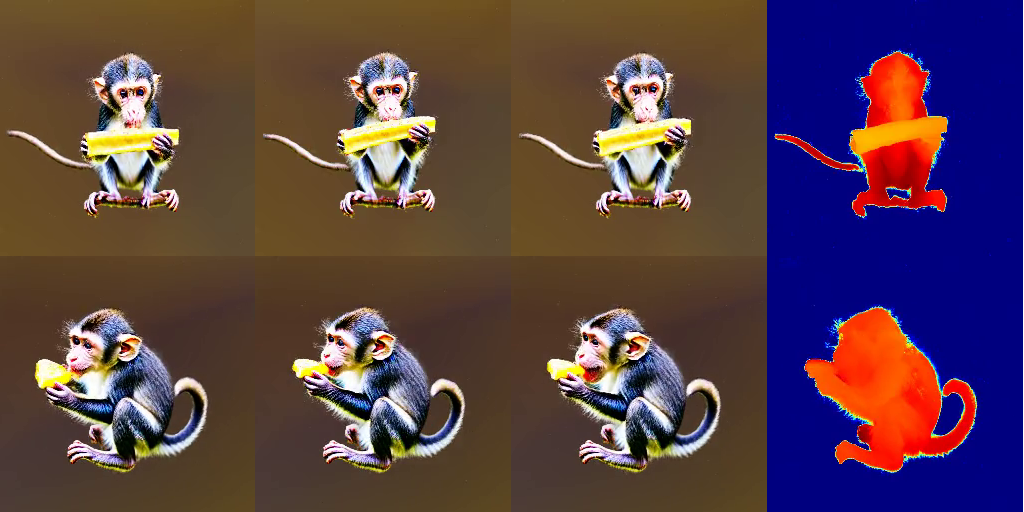}
&\includegraphics[height=\myheight]{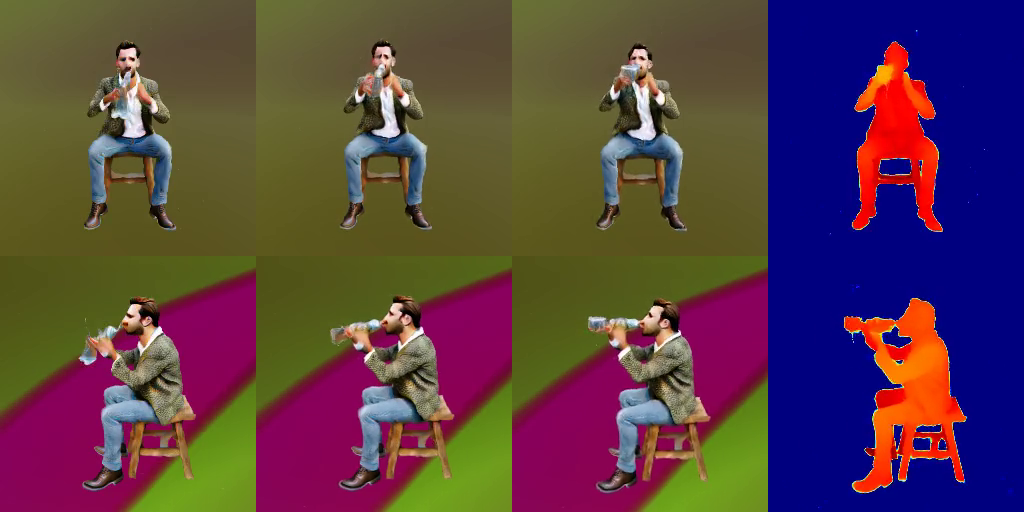}
\\
A monkey is eating a candy bar & A man is drinking beer \\
\includegraphics[height=\myheight]{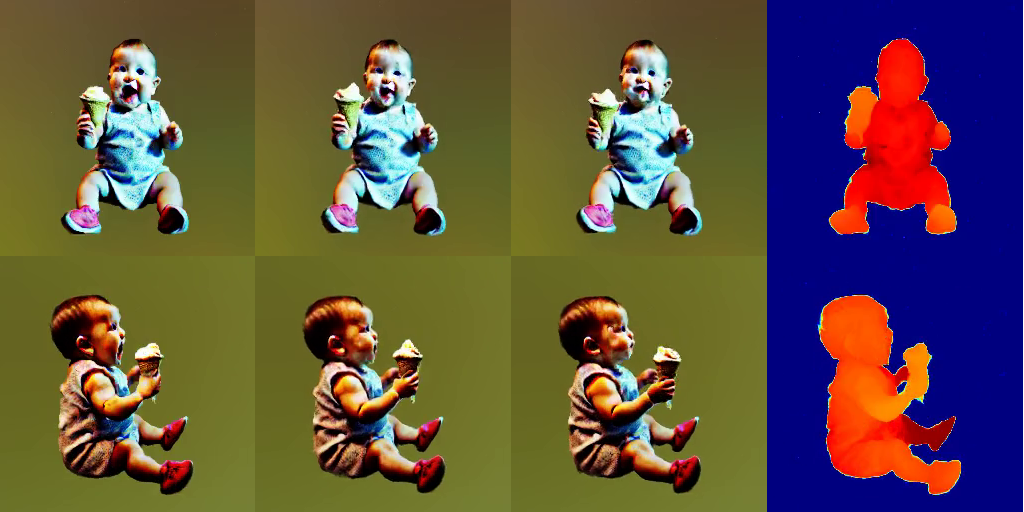}
&\includegraphics[height=\myheight]{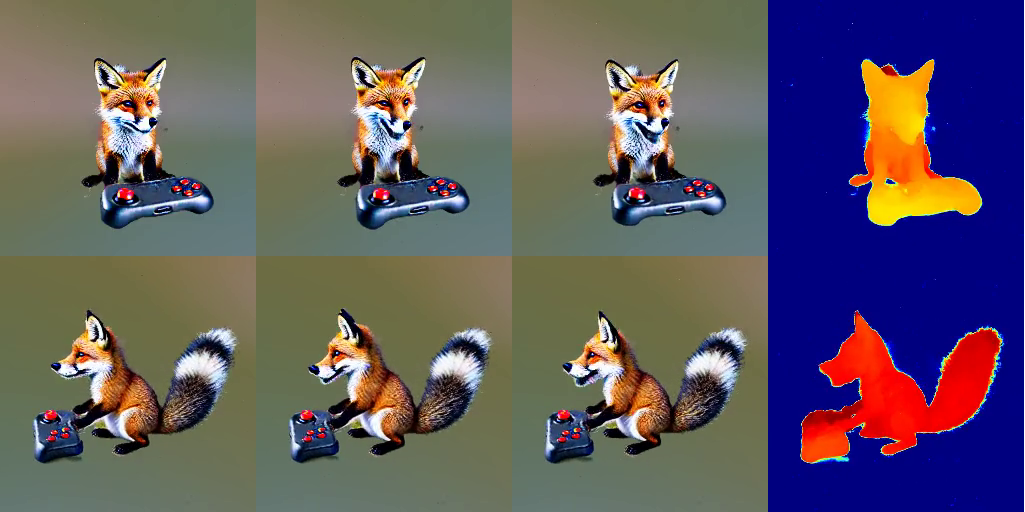}
\\
A baby is eating ice cream & A fox is playing a video game \\
\includegraphics[height=\myheight]{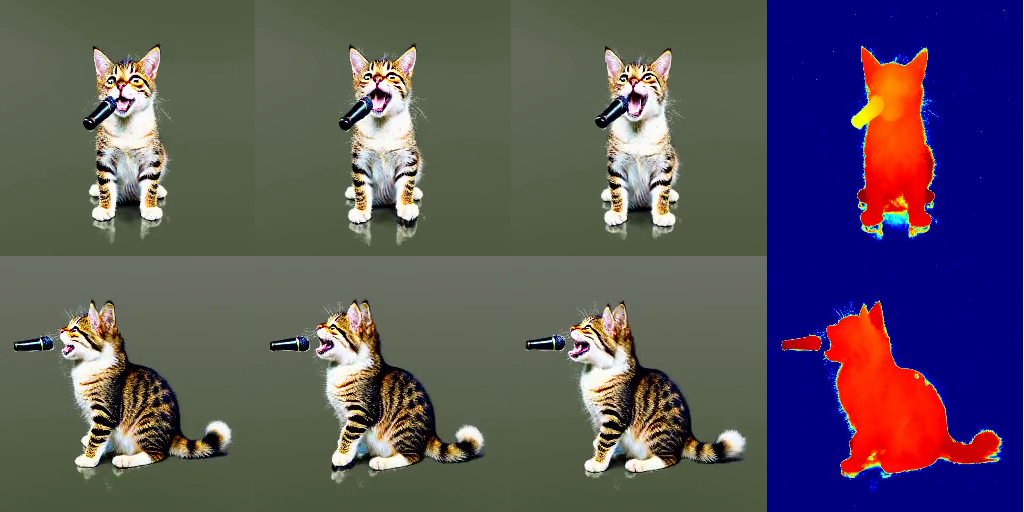}
&\includegraphics[height=\myheight]{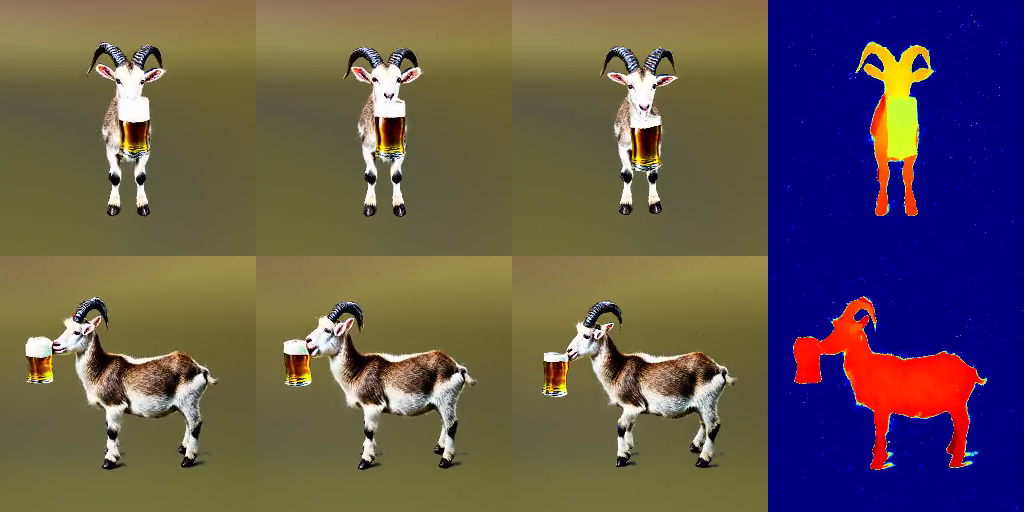}
\\
A cat is singing & A goat is drinking beer \\
\includegraphics[height=\myheight]{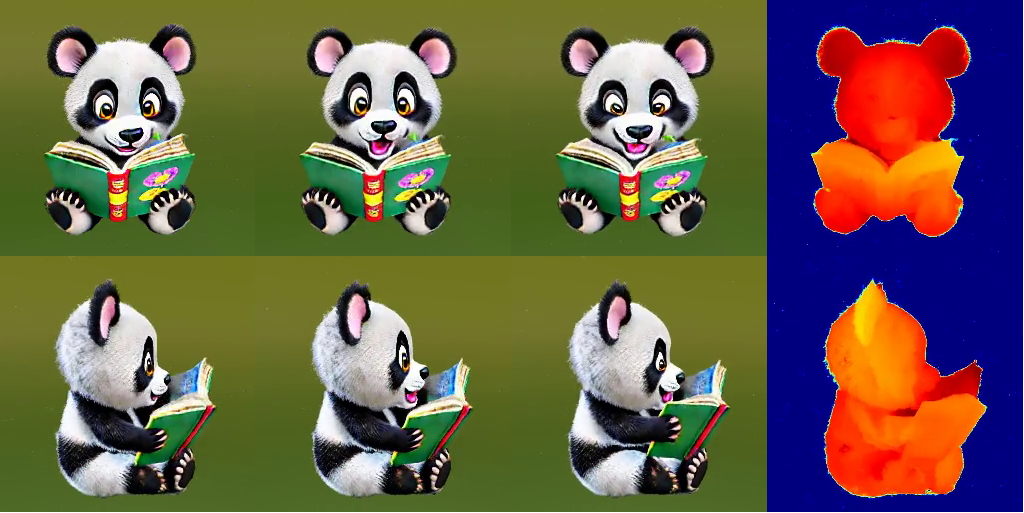}
&\includegraphics[height=\myheight]{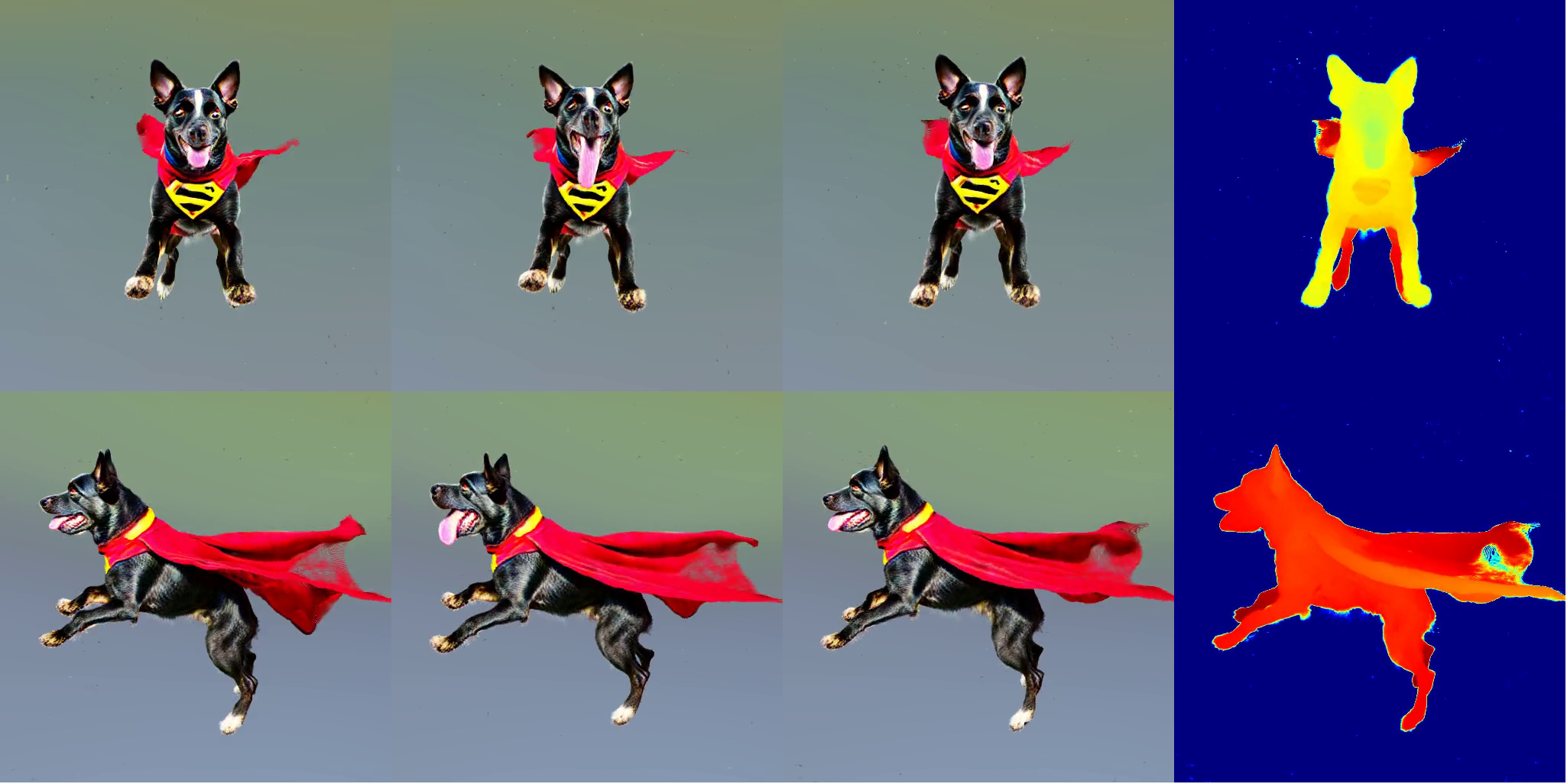}
\\
Emoji of a baby panda reading a book & A superhero dog wearing a red cape is flying through the sky
\end{tabularx}
\caption{
\textbf{Text-to-4D generation. } We show qualitative results of text-to-4D generation, demonstrating high visual quality, multi-view consistency, plausible composition and realistic motion. Video results are available in the \suppmat. 
\label{fig:qualitative_result}
}
\end{center}
\end{figure*}
\section{Results}

\subsection{Text-to-4D Generation}
In Fig.~\ref{fig:qualitative_result}, we show qualitative results of our method on text-to-4D generation. Video results are displayed on the project webpage for better assessment of the motion. 
\vskip 1em
\myparagraph{User study.} We carry out a user preference study to evaluate sample quality along the dimensions of
\begin{inparaenum}[(1)]
\item alignment to the input text prompt, 
\item motion quality, and 
\item 3D consistency and visual quality. 
\end{inparaenum}
For each vote, we present the participant with results from our method as well as from the baseline(s), and ask the participant to pick the best method given one of the three evaluation metrics described above. 
\begin{table}[ht!]
    \vskip -2mm
    \centering
    \begin{tabular}{lccc}
    \toprule
    \small{Metric} & \small{visual \& 3D} & \small{text alignment} & \small{motion}\\
    \midrule
    Ours & 82.4\%& 65.4\%& 61.8\% \\
    \bottomrule
    \end{tabular}
    \vskip -2mm
    \caption{\textbf{Comparison with MAV3D.}  The numbers indicate the percentage of users who prefer our results over MAV3D's.
    }
    \label{tab:MAV3D_comparison}
\end{table}

\vskip 1em
\myparagraph{Comparison with MAV3D. }Since there is no publicly available implementation of MAV3D~\cite{singer2023textto4d}, we compare our method against the 28 visual results displayed on their website. Through a user study with 13 users (results are shown in Tab~\ref{tab:MAV3D_comparison}), our method outperforms MAV3D in all metrics. 

Additionally, we compare with several ablative baselines following a similar user study protocal with 18 users. We detail the ablation study in the following paragraphs.
\begin{table}[h]
    \begin{subtable}[h]{\linewidth}
        \centering
        \begin{tabular}{lcc}
        \toprule
        \small{Metric} & w/o \small{2D guidance} & \small{Ours}\\
        \midrule
        text alignment  &11.67\% & 88.33\% \\
        \bottomrule
        \end{tabular}
        \caption{\textbf{Text alignment ablation.} Without 2D diffusion guidance, the learned 3D asset might fail to generate all the required components of a scene or fail to produce a plausible layout (see Fig.~\ref{fig:wo_SD}).
        }
        \label{tab:wo_SD_ablation}
    \end{subtable}
\vskip 1em
    \begin{subtable}[h]{\linewidth}
        \centering
        \begin{tabular}{lccc}
        \toprule
        \small{Metric} & \small{w/o Multi-res Grid} & \small{w/o TV loss} & \small{Ours}\\
        \midrule
        \small{motion quality} &42.22\% & 2.78\% & 55.00\% \\
        \bottomrule
        \end{tabular}
    
        \caption{\textbf{Motion quality ablation.} Without the multi-resolution feature grid, detailed  local motion cannot be learned. Without the proposed TV loss, the generated motion contains substantial noise (see Fig.~\ref{fig:motion_ablation}). 
        }
        \label{tab:motion_ablation}
     \end{subtable}
\vskip 1em
     \begin{subtable}[h]{\linewidth}
        \centering
        \begin{tabular}{lccc}
        \toprule
        \small{Metric} & \small{w/o 3D guidance} & \small{Hexplane} & \small{Ours}\\
        \midrule
        \small{visual\&3D} &6.12\% & 0.00\% & 93.88\% \\
        \bottomrule
        \end{tabular}
        \caption{\textbf{Visual quality and 3D consistency abation.} Without 3D diffusion guidance in the first stage, the learned 3D assets suffer from the Janus problem (see Fig.~\ref{fig:wo_SD}). With a hexplane 4D representation, the learned high-quality 3D asset
        cannot be preserved in the dynamic stage, leading to lower visual quality and re-appearance of the Janus problem (see Fig.~\ref{fig:hexplane}). 
        }
        \label{tab:quality_and_3D_ablation}
     \end{subtable}
     \vskip -0.5em
     \caption{\textbf{Comparison with ablative baselines. } }%
     \label{tab:user_study}
\end{table}
\myparagraph{3D and 2D diffusion guidance in the static stage.}
Our method combines 3D and 2D diffusion guidance to learn the static model. Fig.~\ref{fig:wo_SD} shows qualitative comparisons, ablating the guidance used in the static stage. Without 2D diffusion guidance, the method often fails to produce the correct layout of the scene, and sometimes produces synthetic-looking texture. This is also reflected by the user study in Tab.~\ref{tab:wo_SD_ablation}. Without 3D diffusion guidance, the learned assets suffer from severe Janus problems and do not have plausible shapes (see row 1 of Fig.~\ref{fig:wo_SD} and `w/o 3D guidance' in Tab.~\ref{tab:quality_and_3D_ablation}).
By combining both 3D and 2D guidance, our method reconstructs 3D-consistent static scenes with plausible compositions and realistic textures.
\\
\myparagraph{Deformation field and motion regularization.}
To learn better motion, we propose to use a multi-resolution hash-encoded 4D feature grid for the deformation MLP. We ablate this choice against a baseline MLP with positional encoding~\cite{mildenhall2021NeRF}. In Fig.~\ref{fig:motion_ablation}, our method learns more local motion around the mouth and fin areas of the clown fish. We also ablate the total variation (TV) loss on the displacement map and show that the learned motion presents substantial noise when not using the TV loss (Fig.~\ref{fig:motion_ablation}).  These observations about the learned motion quality are also reflected by the user study in Tab.~\ref{tab:motion_ablation}.
\\
\myparagraph{D-NeRF v.s. hexplane representation}
We also ablate our deformable NeRF representation against the hexplane 4D representation~\cite{cao2023hexplane} used previously in ~\cite{singer2023textto4d}. Due to the entanglement of geometry and motion in hexplanes, it is not trivial to keep the static geometry parts frozen during the motion learning stage, which leads to lower visual quality and reappearance of the Janus problem (See Fig.~\ref{fig:hexplane} and Tab.~\ref{tab:quality_and_3D_ablation}, `Hexplane' versus Ours). In comparison, our dynamic representation fully disentangles the canonical model and the deformation field, successfully preserving the static model while learning its motion.

\subsection{Controllable 4D Generation}
\begin{figure}[t]
\begin{center}
\includegraphics[trim=3em 0em 0em 0em, clip=true, width=1.0\linewidth]{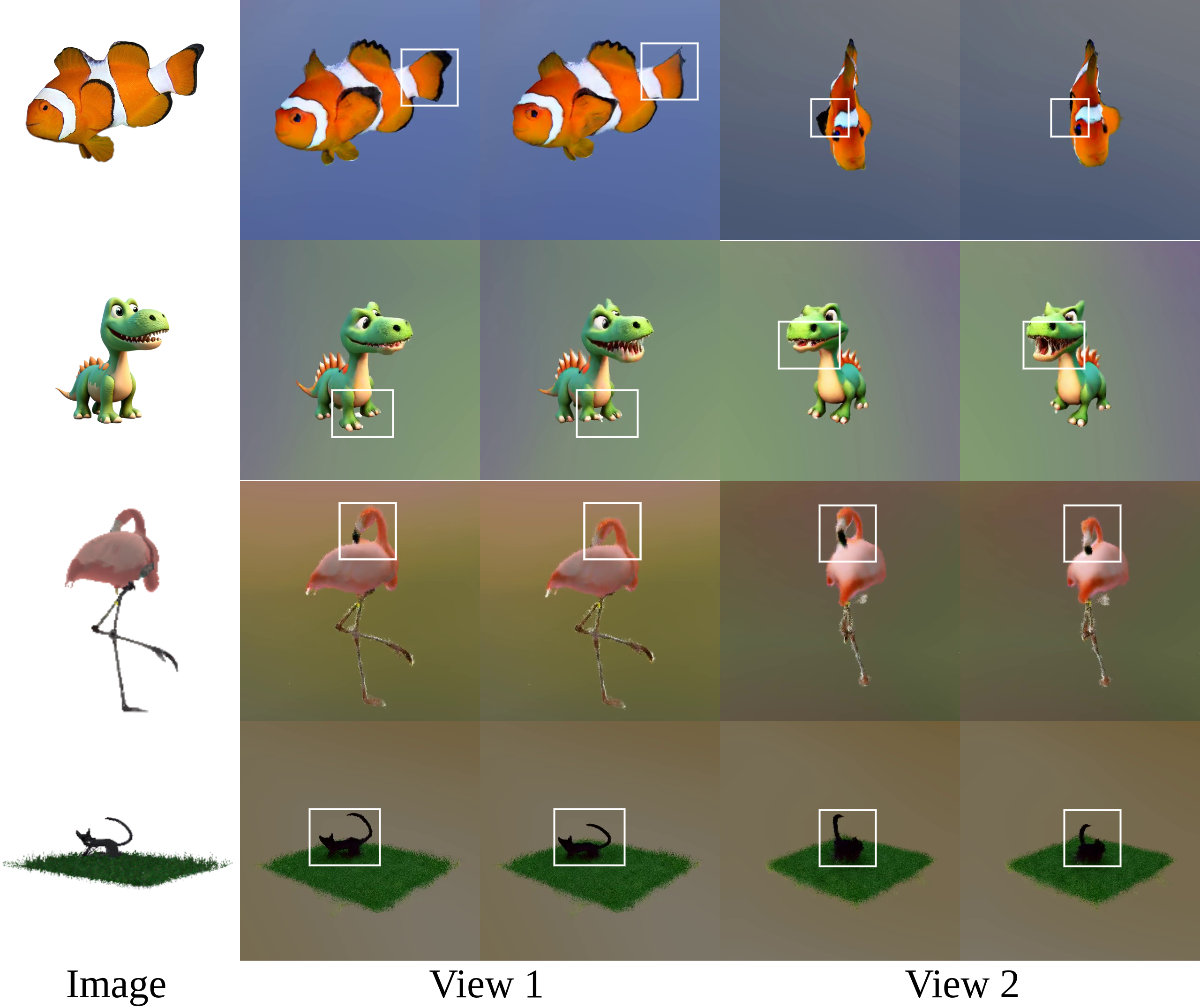}
\end{center}
\vspace{-0.4cm}
\caption{
\textbf{Image-to-4D generation. } Given an input image, our method reconstructs and animates 3D assets. Prompts used for motion are `A clown fish swimming', `A cartoon dragon running', `A flamingo scratching its neck', and `A cat walking on grass'. Video results can be found in our Supplement.}
\label{fig:image-to-4D}
\vspace{-0.4cm}
\end{figure}

\begin{figure}[t]
\begin{center}
\includegraphics[trim=0em 0em 0em 0em, clip=true, width=1.0\linewidth]{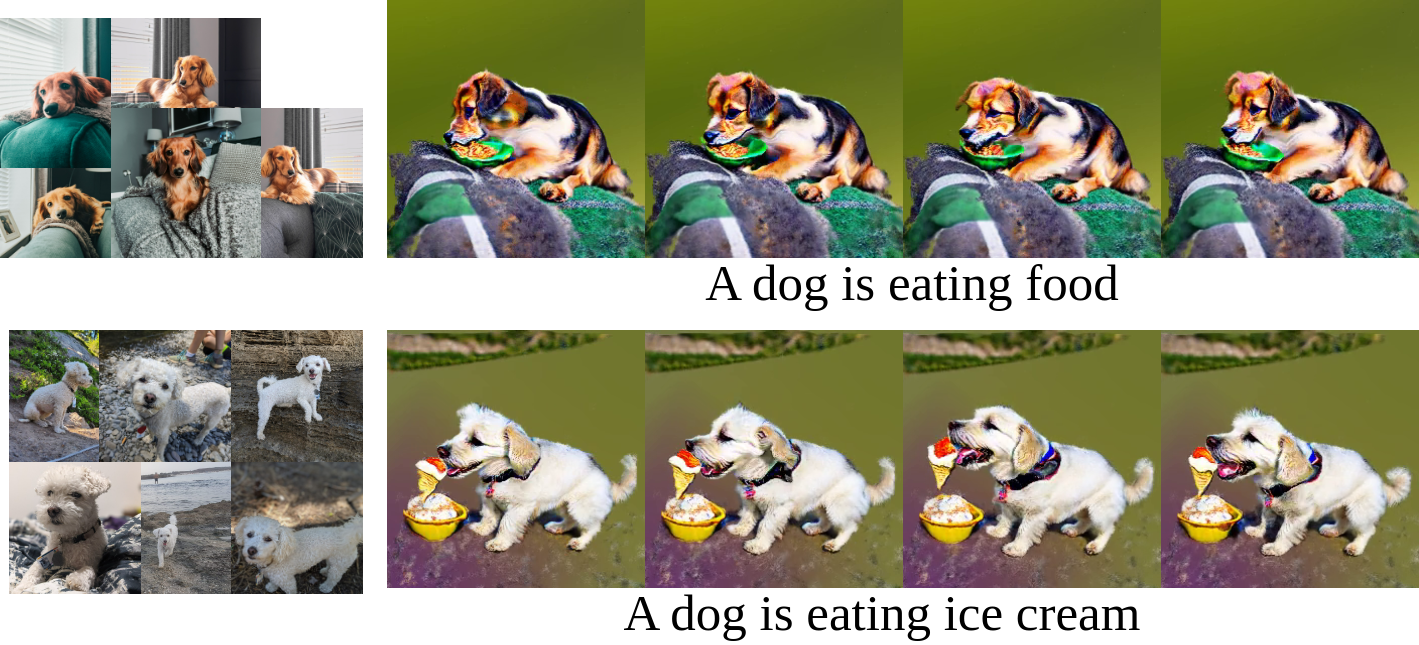}
\end{center}
\vspace{-0.5cm}
\caption{
\textbf{Personalized 4D. } Our method can generate dynamic 3D scenes of a subject given a text prompt and 4-6 causally captured images of the subject. Videos are available in our Supplement. }
\label{fig:personalized_4D}
\end{figure}

We show qualitative results of text-to-4D generation where the object appearance is defined by one or multiple user-defined images. In Fig.~\ref{fig:image-to-4D}, given a single image, our method can preserve the identity and appearance details of the input image and successfully learn the animation specified in the text prompt. It is preferred by users over MAV3D 98.3\% of the time for its image texture preservation and 100.0\% of the time for its alignment to the text. In Fig.~\ref{fig:personalized_4D}, we show personalized 4D generation. Given a few casually captured images of a subject, our method can generate 4D content of the subject under  various motion conditions, \textit{e.g.}, eating food or ice cream. More qualitative results can be found on our project page.

\section{Discussion}
\begin{figure}[t]
\begin{center}
\includegraphics[trim=3em 0em 0em 0em, clip=true, width=1.0\linewidth]{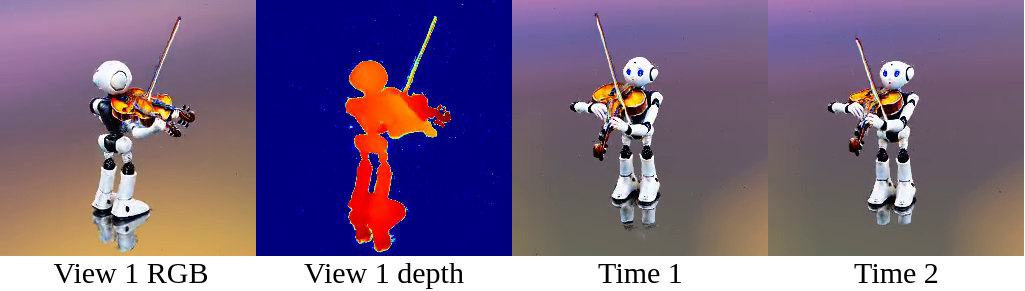}
\end{center}
\vspace{-0.4cm}
\caption{
\textbf{Failure case. } Despite combining 3D and 2D diffusion guidance, our method fails to reconstruct `A robot is playing the violin'. The static stage fails to learn a view-consistent violin, and the robot's hand position is incorrect. In the second stage, our method cannot correct such errors or learn plausible arm motion.}
\label{fig:failure_case}
\vspace{-0.5cm}
\end{figure}

\paragraph{Conclusion} We propose \methodname, a unified approach to 4D scene synthesis from a text prompt and optionally one or more input images. %
By leveraging 3D and 2D diffusion priors, our method first learns a high-quality static asset, offering a good starting point for deformation optimization. Then, our motion-disentangled 4D representation allows us to learn motion with video diffusion guidance while maintaining the quality of the static asset. We introduce multi-resolution feature grids and a TV loss for the deformation field, resulting in more realistic motion. \methodname{} achieves better visual quality, 3D consistency, motion and spatial layout on the text-to-4D task compared to the baselines, while also enabling image-to-4D and personalized 4D generation. 
\\
\myparagraph{Limitation} The combination of 3D and 2D diffusion priors sometimes fails to learn correct static 3D representations for some difficult prompts (\textit{e.g.}, a robot playing a violin in Fig.~\ref{fig:failure_case}). 
In the dynamic stage, our method cannot recover from the wrong static representation and fails to learn correct motion given the wrong position of the hands. 
We believe this problem could potentially be solved with further advances in 3D and 2D diffusion models. 
\\
\myparagraph{Acknowledgement}  We sincerely thank all participants of our user study for their help. Yufeng Zheng is partially supported by the Max Planck ETH Center for Learning Systems. 
\pagebreak

\begin{center}
\textbf{\large Supplemental Materials}
\end{center}
\setcounter{equation}{0}
\setcounter{figure}{0}
\setcounter{table}{0}
\setcounter{page}{1}
\setcounter{section}{0}
\section{Societal impact} 
We note that our work could be potentially used to generate fake 4D content that is violent or harmful. Building upon pre-trained large-scale diffusion models, it inherits the biases and limitation of these models. Therefore, the 4D videos generated with our method should be carefully examined and labeled as synthetic content. 
\section{More Results}
\begin{figure}[h!]
\vspace{-3mm}
\begin{center}
\newcommand{\myheight}{1.5cm}
\setlength\tabcolsep{1pt}
\begin{tabularx}{\textwidth}{cccc}
\includegraphics[trim={0 2cm 0 3cm},clip, height=\myheight]{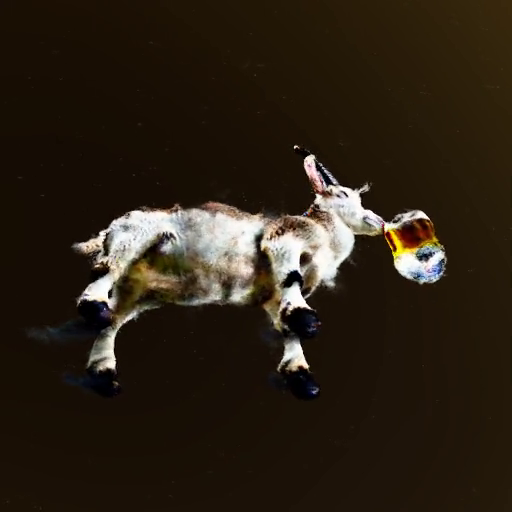}
&\includegraphics[trim={0 2cm 0 3cm},clip, height=\myheight]{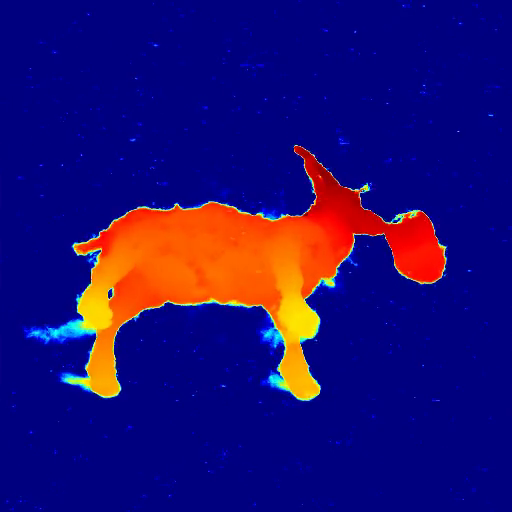}
&\includegraphics[trim={0 3cm 0 2cm},clip, height=\myheight]{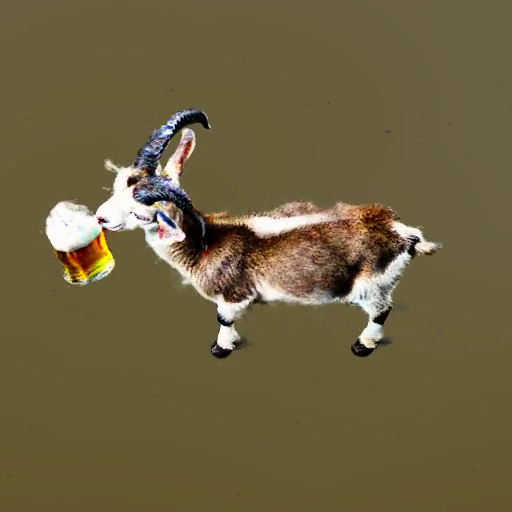}
&\includegraphics[trim={0 3cm 0 2cm},clip, height=\myheight]{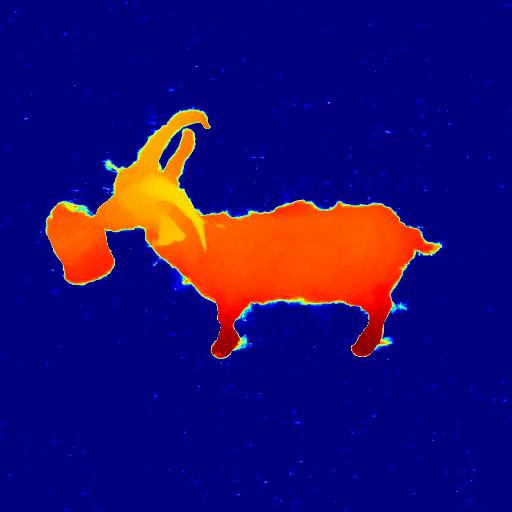}

\end{tabularx}
\vskip -3mm
\caption{
\textbf{Renderings from different elevation angles}.
\label{fig:rebuttal_elevation_angle}
}
\end{center}
\vspace{-1cm}
\end{figure}

\vskip 1mm
\myparagraph{Video results}
Please refer to our web page (\url{https://research.nvidia.com/labs/nxp/dream-in-4d/}) for video results of the text-to-4D, image-to-4D and personalized 4D tasks. We also provide qualitative comparisons of our method and ablation baselines. %
\\
\myparagraph{More rendering angles}
In Fig.~\ref{fig:rebuttal_elevation_angle}, we show renderings from the top view and the bottom view. Our 4D assets produce plausible renderings from different elevation angles. 

\section{Network Architecture}
\vskip 5mm
\myparagraph{Canonical NeRF.}
We use a hash-encoded multi-resolution feature grid with $16$ resolution levels, where the base resolution is $16\times16\times16$ and maximum resolution is $4096\times4096\times4096$. The feature grid is followed by two shallow MLPs to produce the density and color values. Both of the MLP networks have $1$ hidden layer and $64$ neurons per layer. 
\\
\myparagraph{Deformation field.}
The deformation field uses a hash-encoded multi-resolution feature grid with $12$ levels, where the base resolution is $4\times4\times4\times4$ and maximum resolution is $232\times232\times232\times232$. The feature grid is followed by an MLP with $4$ hidden layers and $64$ neurons per layer to predict the displacement values $\mathbf{d}$ for scene deformation. We then calculate the canonical point location by $\mathbf{x_c} = \mathbf{x_d} + \mathbf{d}$.
\\
\myparagraph{Background.}
We model the background with an MLP, which takes the viewing direction as input and outputs a color value. This assumes that the background is located infinitely far away from the camera. The MLP has $3$ hidden layers and $64$ neurons per layer. %
\section{Training Schedule}
\vskip 5mm 
\subsection{Static Stage}
For the static stage, we render multi-view images of resolution $64\times64$ with a batch size of $8$ for the first $5000$ iterations, and resolution $256\times256$ with a batch size of $4$ for the last $5000$ iterations. For MVDream~\cite{shi2023MVDream} guidance, the images are upsampled to $256\times256$. For StableDiffusion~\cite{rombach2021highresolution}, we upsample to $512\times512$. 
We use guidance scale of $50$ for MVDream and $100$ for StableDiffusion. %

We use an AdamW~\cite{loshchilov2017adamw} optimizer with a learning rate of $0.001$ for all the MLP parameters and $0.01$ for the parameters of hash-encoded multi-resolution feature grid. The $\beta$ parameters are set to $0.9$ and $0.99$, respectively. We train the networks for $10000$ iterations on a NVIDIA V100 GPU, which takes $4.5$ hours.

We found that balancing the 3D and 2D guidance weights to be important for achieving view-consistent, text-aligned and realistic results. In Tab.~\ref{tab:loss_weights}, we list the loss weights used for all prompts in the paper. 
\begin{table}
\resizebox{\linewidth}{!}{%
\begin{tabular}{ccc}
\toprule
Prompts & $\lambda_{2D}$ & $\lambda_{3D}$ \\
\midrule
A superhero dog wearing a red cape flying through the sky & 1.2 & 1\\
A cat singing & 1.2 & 1\\
A dog riding a skateboard & 1 & 1\\
Clown fish swimming through coral reef & 1 & 1\\
A fox playing a video game & 1.2 & 1\\
A goat drinking beer & 1 & 1\\
A monkey eating a candy bar & 1 & 1\\
An emoji of a baby panda reading a book & 1.2 & 1\\
A baby panda eating ice cream & 1 & 1\\
A squirrel riding a motorcycle & 1.2 & 1\\
\bottomrule
\end{tabular}%
}
\caption{2D and 3D guidance loss weights for the static stage. }
\label{tab:loss_weights}
\end{table}

\subsection{Dynamic Stage}
To learn deformation in the dynamic stage, we use guidance from the Zeroscope video diffusion model~\cite{zeroscope}, where we render $24$-frame videos of resolution $144\times80$. We found our method to also work well with guidance from the Modelscope video diffusion model~\cite{modelscope}, in which case we rendered videos of resolution $64\times64$ for the first $7000$ iterations and then upsample to $256\times256$. We use a guidance scale of $100$ for both Zeroscope and Modelscope guidance, and gradually decrease the time step used for the SDS loss~\cite{poole2022dreamfusion} from $[0.99, 0.99]$ to $[0.2, 0.5]$. 

In the dynamic stage, we optimize the deformation parameters with an AdamW~\cite{loshchilov2017adamw} optimizer with learning a rate of $0.001$ and $\beta=[0.9, 0.99]$. We train the deformation network for $10000$ iterations on a NVIDIA A100 or RTX A6000 GPU. We start by using the first $4$ levels of the multi-resolution features, and gradually include the higher resolution features, adding $1$ level every $500$ iterations. The dynamic stage takes $9$ hours for Zeroscope, and $6$ hours for Modelscope due to the lower inference resolution of the diffusion model. 

\section{User Study}
\begin{figure}[t]
\begin{center}
\includegraphics[trim=5em 0em 5em 0em, clip=true, width=1.0\linewidth]{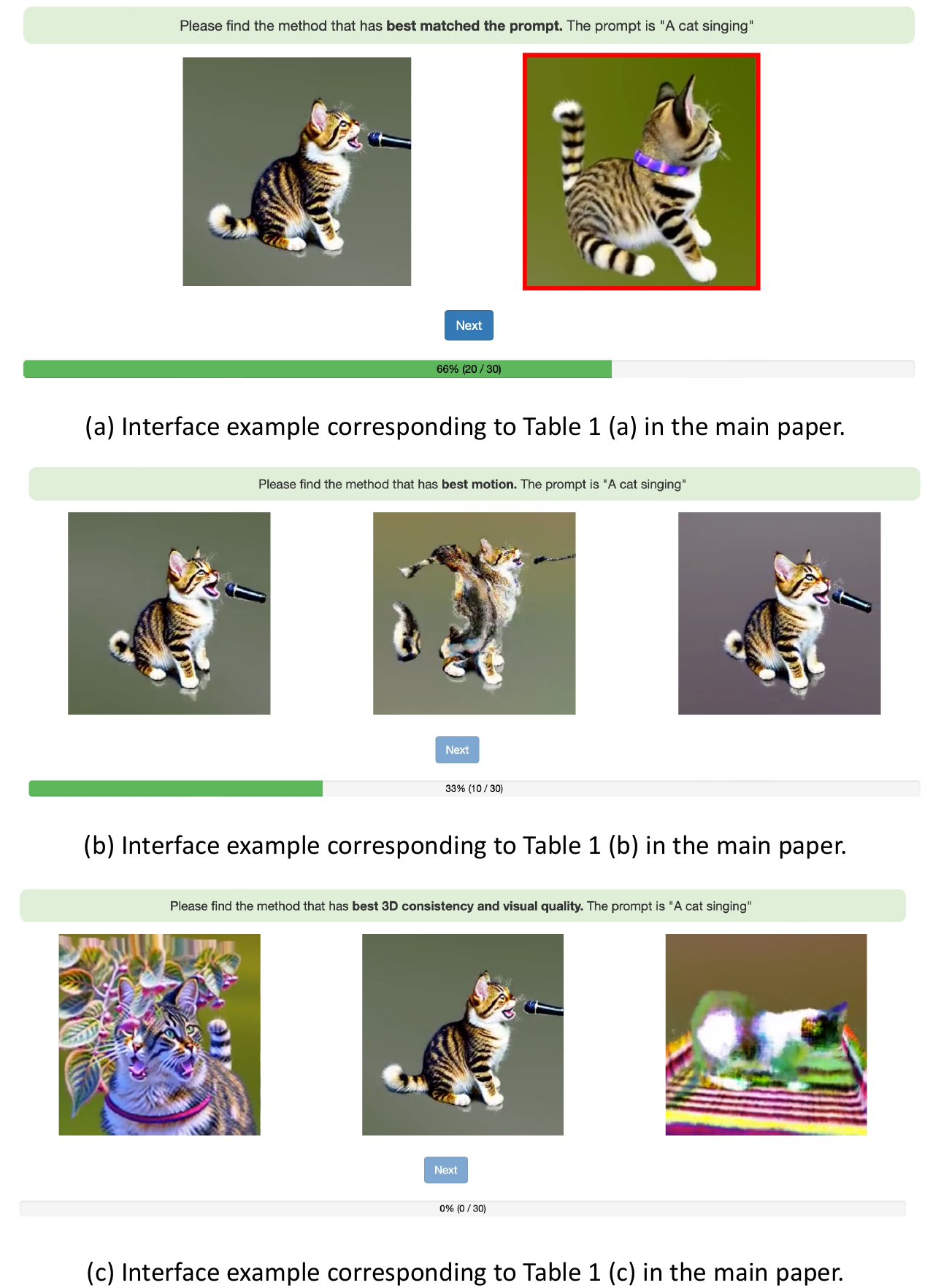}
\end{center}
\caption{
\textbf{User study interface.}}
\label{fig:user_study}
\vspace{-0.5cm}
\end{figure}
In this section, we introduce more details of the user study.
As discussed in Sec.~4.1 of the main paper, we compare against state-of-the-art method (MAV3D) and ablative baselines through a user preference study. 
Specifically, we present the results from our method and the baseline(s)
to a user and give instructions to ``Please find the method that best matched the prompt / has best motion / has best 3D consistency and visual quality."
The example interface is shown in Fig.~\ref{fig:user_study}.

{
    \small
    \bibliographystyle{ieeenat_fullname}
    \bibliography{main}
}

\end{document}